# Preserving Seasonal and Trend Information: A Variational Autoencoder-Latent Space Arithmetic Based Approach for Non-stationary Learning

Hassan Wasswa
*School of Systems and Computing*
*University of New South Wales*
Canberra, Australia
h.wasswa@unsw.edu.au

Aziida Nanyonga
*School of Engineering and Technology*
*University of New South Wales*
Canberra, Australia
a.nanyonga@unsw.edu.au

Timothy Lynar
*School of Systems and Computing*
*University of New South Wales*
Canberra, Australia
t.lynar@unsw.edu.au

*Abstract*—AI models have garnered significant research attention towards predictive task automation. However, a stationary training environment is an underlying assumption for most models and such models simply do not work on non-stationary data since a stationary relationship is learned. The existing solutions propose making data stationary prior to model training and evaluation. This leads to loss of trend and seasonal patterns which are vital components for learning temporal dependencies of the system under study. This research aims to address this limitation by proposing a method for enforcing stationary behaviour within the latent space while preserving trend and seasonal information. The method deploys techniques including Differencing, Time-series decomposition, and Latent Space Arithmetic (LSA), to learn information vital for efficient approximation of trend and seasonal information which is then stored as embeddings within the latent space of a Variational Autoencoder (VAE). The approach's ability to preserve trend and seasonal information was evaluated on two time-series non-stationary datasets. For predictive performance evaluation, four deep learning models were trained on the latent vector representations of the datasets after application of the proposed method and all models produced competitive results in comparison with state-of-the-art techniques using RMSE as the performance metric.

## I. INTRODUCTION

The advancement of computing technologies, like Graphical Processing Units (GPUs), has enabled the implementation of resource-intensive AI algorithms, including deep learning models of complex architecture, across various fields such as aviation [1], security, and data privacy [2]–[4] and many more. Machine learning models typically assume that training and test data share the same distribution, maintaining stationary statistical properties. Under this assumption, models achieve high accuracy and precision in classification and regression tasks [5]. However, many real-world datasets are non-stationary, causing models to generalize poorly to new data. Deploying models trained on stationary data in non-stationary environments can result in misleading outputs with significant consequences [6].

Non-stationary behaviour can arise from legitimate sources like a growing population causing a rising trend in water consumption [7]. Conversely, non-legitimate sources introduce abrupt "spikes" or "dips" in time series data, often detectable as anomalies [8]–[10]. These anomalies are usually triggered by specific, short-lived events, such as a sudden drop in electricity usage due to a malfunctioning transformer, which can be identified through long-term trend analysis [11].

However, legitimate non-stationarity sources challenge machine learning models, as accurate predictions depend on this information. The authors in [12] studied the impact of non-stationarity on the performance of both statistical and machine learning models and found that non-stationarity significantly impacts learning model performance. As a solution, researchers often remove non-stationarity from data before training, consequently losing trend and seasonal information which are crucial for understanding temporal dependencies in non-stationary time-series [13].

To address this challenge, this study proposes an approach for non-stationary removal within the VAE's latent space while preserving trend and seasonal patterns. Using time-series decomposition, the dataset is split into trend, seasonal, and stationary residual components, with seasonal data stored as Embeddings within the latent space. This method is inspired by study [14], where discrete vectors are stored as Embeddings for quantizing latent code. The Key contributions of this study include.

1) The study introduces a hybrid method using Differencing, Time-series decomposition, and LSA for non-stationary removal, while retaining trend and seasonal components.
2) By Differencing within the latent space, we demonstrate an efficient way to control the impact of non-stationarity on the predictive performance of learning models.

The rest of the paper is organized as follows: Section II reviews related work; Section III describes the proposed approach, datasets, and techniques; Section IV presents experimental results; Section V discusses findings; and Section VI concludes the study.

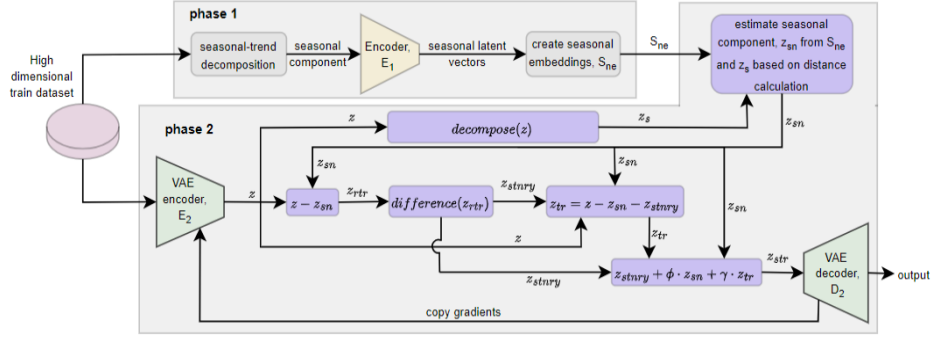

Fig. 1: Proposed non-stationary removal approach

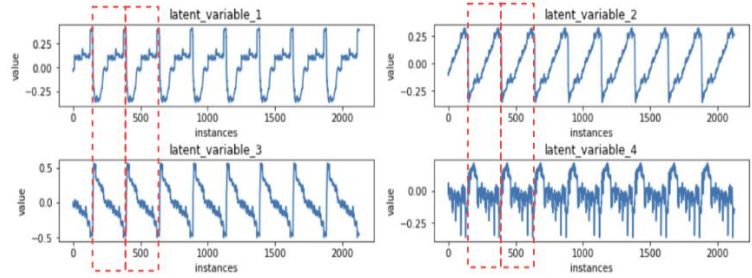

Fig. 2: Seasonal plots of latent variables with red outlines showing repeated period patterns

## II. RELATED WORK

Working with non-stationary data remains challenging for reliable machine learning. Commonly, non-stationary behaviour is removed from the training set before model training and evaluation using techniques like "Detrending" by A.J. Owens (1978) [15], "Filtering" by N.A.A. Abd Rabbo (1966) [16], "Differencing", and "time-series decomposition" by Cleveland and Tiao (1976) [17]. Since, this study used Differencing for non-stationary removal, Time-series decomposition for time-series component extraction, and Deep learning for prediction, the scope of prior work is limited accordingly.

The Differencing technique repeatedly replaces entry $i_t$, at time, $t$, with $i_t - i_{t-1}$ until the series is stationary [18]. Its application is seen in statistical methods like Autoregressive Integrated Moving Average (ARIMA) [19], [20] and Augmented Dickey Fuller (ADF) [21]. However, it degrades performance due to loss of long-term trends [19], [21].

Time-series decomposition, on the other hand, splits a series into its components to allow separate analysis. The Seasonal-Trend decomposition based on Loess (STL) technique was introduced in study [22] to decomposes a time-series into trend, seasonal, and residual components while assuming a simple additive structure and one seasonal pattern.

The study in [23] proposed a seasonal-trend decomposition algorithm that extended the STL-method for robust seasonal-trend decomposition of long-term time-series, accommodating sudden alterations in trend and fluctuations in the series' seasonal patterns.

Study [24] introduced a regression-based decomposition method for series with multiple seasonal patterns, using an additive model for smooth seasonal and trend components. Additionally, study [25] enhanced the RobustSTL scheme [23] to handle complex seasonal, trend, and residual patterns with improved robustness and efficiency.

However, the Differencing method fails to preserve trend/seasonal attributes of the series. Also, while Time-series decomposition allows separate analysis of components, the finding from study [12], indicated that time-series decomposition alone could not benefit machine learning prediction on non-stationary datasets.

## III. PROPOSED APPROACH

This work proposes a two-phase approach for preserving trend and seasonality while suppressing the non-stationarity impact on AI model performance. Fig. 1 illustrates the architectural design of our approach.

In phase 1, a high-dimensional non-stationary dataset is decomposed into seasonal, trend, and residual components using Time-series decomposition, but only the seasonal component is used to train a VAE model in this phase, capturing seasonal patterns in the latent space. Using learned Encoder, $E_1$, seasonal data is projected from space $R^d$ to latent space $R^k$, where $k < d$, and stored as non-learnable Embeddings, $S_{ne}$. Since the seasonal component constitutes a periodic pattern that repeats for the entire observation (See Fig. 2), only information spanning one period $T$, is extracted and stored.

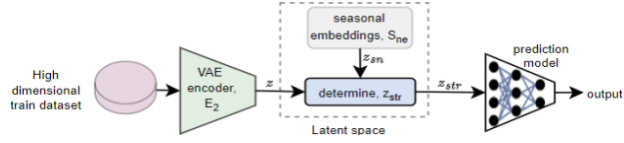

Fig. 3: Training prediction model on stationary low dimensional latent space feature vectors

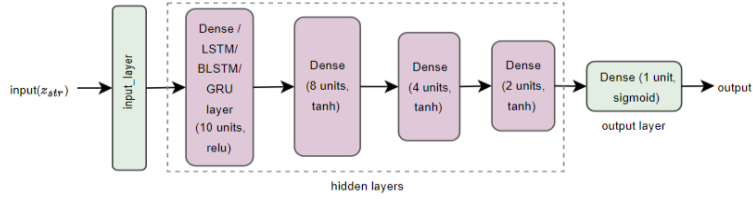

Fig. 4: Prediction model architecture

The main aim of phase 1 is to extract and store a correct approximation of seasonal patterns in vector space, $R^k$.

In phase 2, the VAE encoder, $E_2$, projects the high-dimensional non-stationary dataset to a low-dimensional latent space $z \in R^k$. Time-series decomposition is applied to z to get seasonal latent vectors $z_s$. Through distance calculations [14] between $z_s$ and $S_{ne}$, a refined seasonal component $z_{sn}$ is obtained. Subtracting $z_{sn}$ from $z$ yields $z_{rtr}$, which includes only residual and trend information. Differencing $z_{rtr}$ results in stationary latent codes $z_{stnry}$. And the trend component, $z_{tr}$, can be approximated using Eq. 1.

$$z_{tr} = z - (z_{stnry} + z_{sn}) \quad (1)$$

Finally, through LSA (see Eq. 2), a fraction of seasonal and trend information, $\phi$ and $\gamma$, respectively, is added to the stationary latent series, $z_{stnry}$, obtaining $z_{str}$ which is passed to the VAE-decoder, $D_2$ for reconstruction.

$$z_{str} = v_{stnry} + \phi \cdot z_{sn} + \gamma \cdot z_{tr} \quad (2)$$

where;
$0 \leq \phi \leq 1$ and $0 \leq \gamma \leq 1$ define how much of seasonal and trend information to be preserved, respectively.

### A. Loss function

The loss function accounts for the reconstruction loss, $L_{recon}$, and the stationarization loss, $L_{stnry}$. For an instance $x_i$, let $z(x_i)$ be the encoder output before stationarization and $z(x_i)_{stnry}$ be the stationary latent code. The goal is to minimize $L_{stnry}$, computed as the mean squared error between $z$ and $z_{stnry}$, as in Eq. 3.

$$L_{stnry} = \frac{1}{n}\sum_{i}^{n}(z(x_i) - z(x_i)_{stnry})^2 \quad (3)$$

Therefore, the loss function for this study is defined in Eq. 4:

$$L = L_{recon} + L_{stnry} \quad (4)$$

### B. Dataset

Two datasets were used: the DJIA Dataset[1], with over *5,000* records from *1st January 2000* to *31st December, 2022*, with a total of *7* variables (a Date, and 6 numerical columns); and the NIFTY-50 Stock Market Data[2], including *50* stock categories with over *4,900* data points each, and *15* attributes (a Date, *2* categorical, *12* numerical columns).

TABLE I: VAE encoder and decoder summary

| VAE Encoder | | | | VAE Decoder | | | |
|---|---|---|---|---|---|---|---|
| Layer | Filters/ units | Kernel size | strides | Layer | Filters/ units | Kernel size | strides |
| Input | - | - | - | Dense | 4 | 2 | 3 |
| Conv1D | 32 | (2,2) | 1 | reshape | - | - | - |
| Batch norm + leakyReLU | | | | Batch norm + leakyReLU | | | |
| Conv1D | 16 | (2,2) | 2 | Conv1DT | 8 | (2,2) | 2 |
| Batch norm + leakyReLU | | | | Batch norm + leakyReLU | | | |
| Conv1D | 8 | (2,2) | 2 | Conv1DT | 16 | (2,2) | 1 |
| Batch norm + leakyReLU | | | | Batch norm + leakyReLU | | | |
| Conv1D | 4 | (1,1) | 1 | Conv1DT | 1 | (2,2) | 1 |
| Batch norm + leakyReLU | | | | Batch norm + leakyReLU | | | |
| Flatten | - | - | - | Conv1DT | 1 | (2,2) | 1 |
| Dense | 4 | - | - | output | - | - | - |
| leakyReLU | | | | leakyReLU | | | |

### C. Experiments

VAEs in both phases constituted the same structure. The Encoder included an input layer, four Conv1D layers, each followed by a batch-normalization and a leakyReLU layer, followed by flatten and dense layers. The Decoder used Conv1DTranspose (represented as "Conv1DT" in Table I) layers, each followed by batch-normalization and leakyReLU. The model was trained with Adam optimizer, learning rate *1e-4*, batch size *64*, epochs *30*, and validation split *0.2*, as summarized in Table I.

---
[1] https://www.investing.com/indices/us-30-historical-data
[2] https://www.kaggle.com/datasets/rohanrao/nifty50-stock-market-data

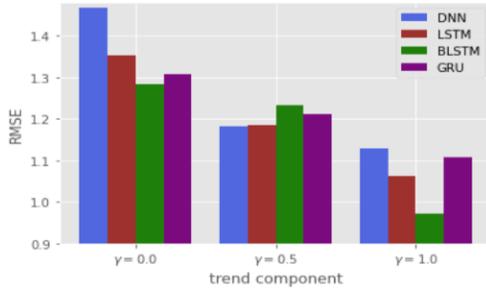
(a) DJIA dataset

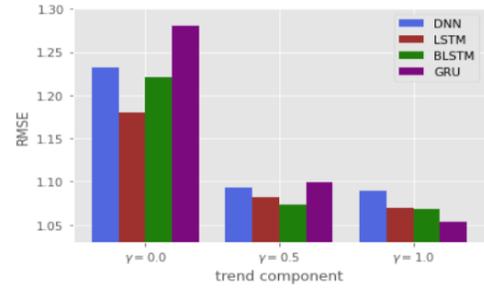
(b) NIFTY-50 Stock Market

Fig. 5: Visualizing the impact of the trend component on model performance of a Time-series

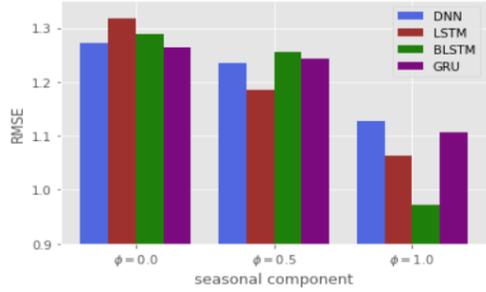
(a) DJIA dataset

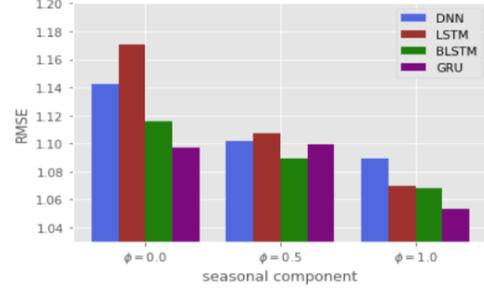
(b) NIFTY-50 Stock Market

Fig. 6: Visualizing the impact of seasonal information on model performance of a Time-series

The datasets were tested for non-stationarity with the ADF test in the high-dimensional space. A standard VAE with latent space dimension 4 was trained, datasets were projected to the latent space and retested for non-stationarity. A VAE was then trained using the proposed approach, datasets projected to the latent space and again tested for non-stationarity. The values of $\phi$ and $\gamma$ were set to 1.0, preserving all seasonal and trend information.

### D. Learning models

To further evaluate the proposed approach, four deep learning models including the standard feed-forward deep neural network (DNN) with back propagation, Long Short-Term Memory (LSTM), Bidirectional Long Short-Term Memory (BLSTM), and Gated Recurrent Units (GRU) were trained, and their performance evaluated.

1) The DNN model: The architecture included an input layer, four hidden dense layers, and an output layer (Fig. 4). ReLU and Sigmoid were used for the first hidden and output layers, respectively, while Hyperbolic Tangent (tanh) was used for all other hidden layers. Training is based on latent space features after LSA (Eq. 2).

2) The LSTM: LSTM, a sophisticated RNN model, incor-porates memory cells and gates (input, output, and forget) to control information flow [26]. The input gate decides what information to store, the output gate handles what to output, and the forget gate manages what to discard. In this study, the LSTM architecture was implemented by replacing the first hidden dense layer in Fig. 4 with a tensorflow-keras LSTM layer, while retaining other dense layers.

3) The BLSTM model: This is an enhancement of the conventional LSTM architecture that processes data bidirectionally, capturing past and future contexts [27]. Its architecture was implemented by replacing the first hidden dense layer of the architecture in Fig. 4 with a Bidirectional-LSTM layer.

4) The GRU: The GRU model addresses the issues of vanishing gradients and computational cost found in LSTMs by using a simpler architecture with Reset and Update gates. The Update gate controls how much past information to retain, while the Reset gate decides what to forget [28]. The Update gate is defined as $u_t = \sigma(W_u \cdot [h_{t-1}, x_t])$, while the Reset gate is given by $r_t = \sigma(W_r \cdot [h_{t-1}, x_t])$, where $u_t$, $r_t$ and $x_t$ are the update, reset gate output and input at time step t, $h_{t-1}$ is the previous hidden state, and $W_u$ and $W_r$ are the weight matrices for the update and reset gate, respectively. The model was implemented by replacing the first hidden dense layer of the architecture in Fig. 4 with a GRU layer while keeping the rest of the hidden layers as dense layers.

## IV. RESULTS

This section presents the experimental findings of this study before and after the application of the proposed approach on each of the two datasets. Each of the datasets was first tested for stationarity on its full dimension before being projected to a 4-dimensional latent space.

A. ADF-Test results from the full dimension datasets

Tables II and III show the ADF test results from the full dimensional DJIA and NIFTY-50 datasets respectively. Table II, reveals that 5 of the 6 numeric features exhibit unit roots with p-values far greater than 0.05. For the NIFTY-50 dataset, as shown in Table III, the p-values of seven fields are far greater than 0.05 implying unit roots exist, hence non-stationary. Also, the corresponding ADF-Statistic values confirm the results with values far greater than the critical value at 1%.

TABLE II: ADF test results from the full dimensional DJIA dataset

| Variable | adf-Statistic | p-value | 1% Critical value |
|---|---|---|---|
| Price | -1.32 | 0.618 | -3.433 |
| Open | -1.49 | 0.537 | -3.433 |
| High | -1.43 | 0.564 | -3.433 |
| Low | -1.46 | 0.550 | -3.433 |
| Vol. | -1.92 | 0.320 | -3.433 |
| Change % | -24.54 | 0.000 | -3.433 |

TABLE III: ADF test results from the full dimensional NIFTY-50 dataset

| Variable | adf-Statistic | p-value | 1% Critical value |
|---|---|---|---|
| Prev Close | -1.46 | 0.553 | -3.44 |
| Open | -1.47 | 0.546 | -3.44 |
| High | -1.44 | 0.565 | -3.44 |
| Low | -1.43 | 0.566 | -3.44 |
| Last | -1.43 | 0.567 | -3.44 |
| Close | -1.42 | 0.568 | -3.44 |
| VWAP | -1.42 | 0.574 | -3.44 |
| Volume | -3.69 | 0.004 | -3.44 |
| Turnover | -11.22 | 0.000 | -3.44 |
| Trades | -6.81 | 0.000 | -3.44 |
| Deliverable Volume | -3.17 | 0.022 | -3.44 |
| %Deliverable | -9.33 | 0.000 | -3.44 |

B. ADF-Test results from low dimensional latent dataset before application of the proposed approach

TABLE IV: ADF Test results on low-dimensional latent space vectors before application of the proposed approach

| Variable | DJIA adf-val | DJIA p-val | NIFTY-50 adf-val | NIFTY-50 p-val |
|---|---|---|---|---|
| l_var1 | -1.01 | 0.749 | -2.19 | 0.210 |
| l_var2 | -1.89 | 0.332 | -1.45 | 0.575 |
| l_var3 | -3.26 | 0.017 | -2.06 | 0.261 |
| l_var4 | -1.79 | 0.386 | -3.91 | 0.002 |

Each dataset is projected to a latent space, $R^k$, (where k = 4), using the encoder model of a standard VAE. The ADF-test results are presented in Table IV. Since the encoding phase does not retain variable names, latent-space variables are referenced by dummy names, prefixed with "$l\_$". The results indicate that projecting a dataset to a low dimensional space does not always guarantee a stationary outcome.

C. ADF-Test results from low dimensional latent dataset after application of the proposed approach

Table V shows the test results in terms of ADF-statistic value (adf-val) and p-value (p-val). The values show that our approach effectively stationarizes all datasets evidenced by all the p-values being less than 0.05.

TABLE V: ADF Test results on low-dimensional latent space vectors after application of the proposed approach

| Variable | DJIA adf-val | DJIA p-val | NIFTY-50 adf-val | NIFTY-50 p-val |
|---|---|---|---|---|
| l_var1 | -9.85 | 0.000 | -48.38 | 0.000 |
| l_var2 | -12.63 | 0.000 | -45.37 | 0.000 |
| l_var3 | -14.09 | 0.000 | -19.76 | 0.000 |
| l_var4 | -10.57 | 0.000 | -46.04 | 0.000 |

TABLE VI: Impact of Trend on the predictive performance of models in terms of RMSE

| Model | DJIA Dataset $\gamma = 0.0$ | 0.5 | 1.0 | NIFTY-50 Dataset 0.0 | 0.5 | 1.0 |
|---|---|---|---|---|---|---|
| DNN | 1.468 | 1.181 | 1.127 | 1.232 | 1.092 | 1.089 |
| LSTM | 1.352 | 1.185 | 1.062 | 1.180 | 1.081 | 1.070 |
| BLSTM | 1.284 | 1.234 | 0.972 | 1.220 | 1.072 | 1.068 |
| GRU | 1.308 | 1.212 | 1.107 | 1.281 | 1.099 | 1.053 |

TABLE VII: Impact of seasonal on the predictive performance of models in terms of RMSE

| Model | DJIA Dataset $\phi = 0.0$ | 0.5 | 1.0 | NIFTY-50 Dataset 0.0 | 0.5 | 1.0 |
|---|---|---|---|---|---|---|
| DNN | 1.2719 | 1.2343 | 1.1276 | 1.1427 | 1.1021 | 1.0891 |
| LSTM | 1.3176 | 1.1853 | 1.0629 | 1.1702 | 1.1072 | 1.0701 |
| BLSTM | 1.2887 | 1.2560 | 0.9721 | 1.1156 | 1.0895 | 1.0681 |
| GRU | 1.2635 | 1.2421 | 1.1071 | 1.0969 | 1.0998 | 1.0533 |

D. Impact of Trend and Seasonal data on learning model performance

1) *Impact of Trend data*: To analyse the impact of trend, the value of $\phi$ was fixed to *1.0* and while $\gamma$ was set to *0.0*, *0.5*, and *1.0*. The results from the DJIA dataset and NIFTY-50 Stock Market Data are shown in Table VI and visualized in Fig. 5a and Fig. 5b.

For the DJIA dataset, models were trained to predict "*Price*" while "*%Deliverable*" was the target variable in the case of NIFTY-50 dataset. During model training, entries from the most recent year were set aside for model testing.

2) *Impact of Seasonal data*: In a similar way, to analyse the impact of seasonal data, $\gamma$ was set to *1.0* while varying $\phi$ to *0.0*, *0.5*, and *1.0*. The results from the two datasets are shown in Table VII and visualized in Fig. 6a and Fig. 6b.

V. DISCUSSION

This work aimed at enhancing the predictive performance of learning models for non-stationary datasets while preserving seasonal and trend information. The proposed method allows the VAE model's encoder to produce a stationary, low-dimensional dataset, even with *100%* trend and seasonal data added (see Table V). Using LSA, the model maintains temporal patterns within the latent space, allowing for effective training or reconstruction of the full dataset.

TABLE VIII: Performance comparison with state-of-the-art methods in terms of RMSE

| Study | Methods (Best model) | Task | RMSE |
|---|---|---|---|
| [29] | Time-series decomposition (CNN-LSTM) | Prediction of NDVI | 0.0573 |
| [30] | Smoothing and Differencing (CNN-LSTM) | Classification and regression for different domains | 1.82 (smallest) |
| [31] | Seasonal-trend decomposition (GRU-based networks) | Rainfall prediction | 7.8895 (overall) |
| Proposed | Time-series decomposition, Differencing, LSA (BLSTM) | %Deliverables prediction | 0.9721 (smallest) |

The predictive performance of four deep learning models was assessed by varying seasonal and trend components, as shown in Fig. 5 and Fig. 6. Results indicate that omitting either component degrades model performance, with RMSE decreasing as more components are included. Also, comparison with state-of-the-art methods using the RMSE revealed that the proposed approach performs competitively on unseen data, as detailed in Table VIII.

## VI. CONCLUSION

Removing non-stationary behaviour from the train dataset has often been considered as a go-to approach for enhancing predictive model efficiency. However, this often results in loss of trend and seasonal patterns which harbour information about temporal dependencies. The work in this study aimed to reduce the impact of non-stationary behaviour while preserving seasonal and trend components.